\documentclass{article}

    \usepackage[preprint, nonatbib]{neurips_2024}

\usepackage[utf8]{inputenc} 
\usepackage[T1]{fontenc}    
\usepackage{url}            
\usepackage{booktabs}       
\usepackage{amsfonts}       
\usepackage{nicefrac}       
\usepackage{microtype}      
\usepackage{xcolor}         

\usepackage{tabularx}
\usepackage{rotating} 

\usepackage{adjustbox}
\usepackage{pifont}
\usepackage{makecell}

\usepackage{times}
\usepackage{soul}
\usepackage{enumerate}
\usepackage{dsfont}
\usepackage[small]{caption}
\usepackage{graphicx}
\usepackage{amsmath}
\usepackage{amsthm}
\usepackage{booktabs}
\usepackage{algorithm}
\usepackage{algorithmic}
\usepackage{arydshln}
\usepackage{multirow}
\usepackage[switch]{lineno}
\usepackage{amssymb}
\usepackage{wrapfig}

\usepackage{color, colortbl}
\definecolor{citecolor}{HTML}{2980b9}
\definecolor{linkcolor}{HTML}{c0392b}

\usepackage[hidelinks,breaklinks=true,colorlinks,bookmarks=false,citecolor=citecolor,linkcolor=linkcolor]{hyperref}

\title{AdapNet: Adaptive Noise-Based Network for Low-Quality Image Retrieval}
\author{%
  ~~ Sihe Zhang$^1$\footnotemark[1]
  ~~ Qingdong He$^2$\footnotemark[1]
  ~~ Jinlong Peng$^2$\thanks{Equal contributions.}
  ~~ Yuxi Li$^2$ 
  ~~ Zhengkai Jiang$^2$ \\
  ~~  \textbf{Jiafu Wu}$^2$ 
  ~~ \textbf{Mingmin Chi}$^1$
  ~~ \textbf{Yabiao Wang}$^2$
  ~~ \textbf{Chengjie Wang}$^2$ \\
  \normalsize $^1$Fudan University ~~ $^2$Youtu Lab, Tencent\\
}

\begin{document}

\maketitle

\begin{abstract}
Image retrieval aims to identify visually similar images within a database using a given query image. Traditional methods typically employ both global and local features extracted from images for matching, and may also apply re-ranking techniques to enhance accuracy. However, these methods often fail to account for the noise present in query images, which can stem from natural or human-induced factors, thereby negatively impacting retrieval performance. To mitigate this issue, we introduce a novel setting for low-quality image retrieval, and propose an Adaptive Noise-Based Network (AdapNet) to learn robust abstract representations. Specifically, we devise a quality compensation block trained to compensate for various low-quality factors in input images. Besides, we introduce an innovative adaptive noise-based loss function, which dynamically adjusts its focus on the gradient in accordance with image quality, thereby augmenting the learning of unknown noisy samples during training and enhancing intra-class compactness. To assess the performance, we construct two datasets with low-quality queries, which is built by applying various types of noise on clean query images on the standard Revisited Oxford and Revisited Paris datasets. Comprehensive experimental results illustrate that AdapNet surpasses state-of-the-art methods on the Noise Revisited Oxford and Noise Revisited Paris benchmarks, while maintaining competitive performance on high-quality datasets. The code and constructed datasets will be made available.
\end{abstract}

\section{Introduction}
The objective of instance-level image retrieval is to search for and retrieve images from a large scale dataset containing the same object as depicted in a given query image. Over the past two decades, various handcrafted feature-based methods~\cite{lowe2004distinctive,zheng2009tour} have been proposed to enhance the performance of instance level image retrieval. Recently, with advancements in deep learning technologies, deep feature representations have emerged as the dominant approach~\cite{noh2017large}. The development of deep neural networks enables the extraction of high-level features that capture complex patterns and semantics, leading to improved retrieval performance~\cite{arandjelovic2016netvlad,cao2020unifying,tolias2020learning}.

However, the efficiency of image retrieval can be severely affected by the quality of query, which is influenced by factors such as resolution, lighting, noise, and compression artifacts~\cite{lin2011perceptual}. A straightforward solution is to include a portion of the noisy dataset as part of the training dataset, but this potentially leads reliance on less reliable visual cues, such as color or textures, to compensate for lost identity information, resulting in inaccurate results~\cite{huang2020improving}. This problem can also be partially addressed through the use of diverse techniques, including denoising, super-resolution, and image enhancement, which belong to low level improvements at the image level~\cite{nachaoui2021regularization}. 
Recent studies start to focus on the training process and aim to guide the model to learn 
\begin{wrapfigure}{r}{0.5\textwidth}
    \centering
    \includegraphics[width=1\linewidth]{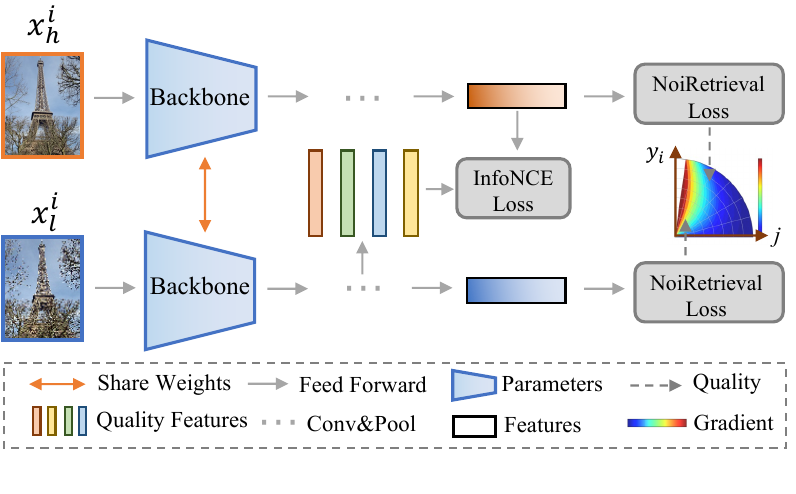}
        \caption{Overview of our proposed method. The current high-quality image input $x_{h}^{i} $ and low-quality image input $x_{l}^{i} $ are fed into the backbone to generate corresponding embeddings. Quality compensation features in multiple colors are employed to learn known noise, and NoiRetrieval Loss is utilized to dynamically adjust the gradient of unknown noise according to image quality.}
	\label{fig:overview}
\vspace{-15pt}
\end{wrapfigure}
robust representation across different image qualities while keeping the discrimination ability. AdaFace~\cite{kim2022adaface} typically proposes a new adaptive margin loss function that emphasizes samples of varying difficulties based on their image qualities. These studies have demonstrated some improvements in recognition accuracy~\cite{wu2022learning}. Nevertheless, AdaFace is trained by sacrificing certain noise data, which may result in relatively weaker discriminative representation.

With the consideration above, in this paper, we propose a novel setting for low-quality image retrieval to retrieve normal images from a database when given low-quality queries. To achieve this, we construct two new noisy datasets by adding various types of random noise to the original test datasets, Revisited Oxford and Revisited Paris~\cite{radenovic2018revisiting}. The noisy dataset is specifically designed to simulate both natural or artificial factors degrading image quality. Judging from partial experimental results, conventional methods struggles in our new benchmarks, making it necessary to develop more robust and adaptive models to handle the disparity in image quality. To address this new problem, we aim to design a noise-based network, as shown in Figure~\ref{fig:overview}, to learn robust abstract representations in low quality images. Specifically, we propose a quality compensation block that leverages high-quality reference images to facilitate the model's understanding of various known noise characteristics. Furthermore, we introduce a novel adaptive noise-based loss function, NoiRetrieval Loss, that dynamically adjusts its attention to gradients based on the quality of images. This adaptive mechanism allows model to enhance unknown noise learning during training and promote intra-class compactness. 
Our main contributions are summarized as follows:
\begin{itemize}
    \item We introduce a novel setting for low-quality image retrieval and propose an adaptive noise-based network (AdapNet) to learn robust abstract representations in low-quality images.
    \item We design a quality compensation block that utilizes high-quality reference images to enhance model's comprehension of different known noise characteristics.
    \item To enhance the model's denoising capability against unknown noise, we design a noise-based loss named NoiRetrieval Loss, which pays more attention to the learning of low-quality samples.
    \item Through extensive experiments, the proposed method achieves stage-of-the-art low-quality image retrieval performance on our new benchmarks: Noise ROxf (+1M), Noise RPar (+1M) while maintaining competitive performance on the original high-quality datasets.

\end{itemize}

\section{Related Work}
\subsection{Image Retrieval}
In earlier studies, techniques such as Fisher vector~\cite{jegou2011aggregating}, VLAD~\cite{jegou2010aggregating}, or ASMK~\cite{tolias2016image} were used to develop global features by aggregating hand-crafted local features. Subsequently, spatial verification methods, such as RANSAC~\cite{fischler1981random}, were employed to refine the initial retrieval results by performing local feature matching. Recently, handcrafted features have been replaced by global and local features extracted from deep learning networks. Notable advancements have been made by leveraging discriminative geometry information in studies~\cite{zheng2017sift,he2018local,dusmanu2019d2,weinzaepfel2022learning}. CVNet~\cite{lee2022correlation} use curriculum learning with the hard negative mining and Hide-and-Seek strategy to handle hard samples without losing generality. The state-of-the-art approach CFCD~\cite{zhu2023coarse} attentively selects prominent local descriptors and infuses fine-grained semantic relations into the global representation. SENet~\cite{lee2023revisiting} captures the internal structure of the images and gradually compresses them into dense self-similarity descriptors while learning diverse structures from various images. Despite the varying quality of images in the dataset, these approaches overlook the impact of noise on image retrieval performance. Consequently, the inclusion of low-quality query images can significantly reduce the overall retrieval effectiveness~\cite{zheng2019uncertainty}. Taking this into consideration, we introduce a training approach that uses pairs of high and low-quality images, deliberately introduces noise, and proposes a quality compensation block to improve the model's understanding of various known noise characteristics.

\subsection{Recognition with Low-Quality Images}
Currently, the work on low-quality image recognition or retrieval is particularly prevalent in the field of face recognition, primarily due to the availability of naturally noisy datasets. Optimal performance depends on two key aspects:1) mastering the extraction of discriminative features from low-quality images, and 2) developing the ability to discard images that contain few identity cues, a process often termed as quality-aware fusion~\cite{kim2022adaface}. To improve quality-aware fusion, many studies have achieved high performance to predict uncertainty representation~\cite{chang2020data,li2021spherical,shi2019probabilistic}
. While AdaFace~\cite{kim2022adaface} has shown effectiveness in face recognition tasks by incorporating image quality to dynamically update the model's gradient weights. It is designed to disregard those low-quality images that are difficult to recognize rather than emphasize learning from low-quality images. Face recognition primarily focuses on identifying or verifying the identity of individuals in images or videos. It grapples with challenges such as variations in facial expressions, poses and other natural noises~\cite{li2023unitsface}. The image retrieval task also faces similar noise issues as the face recognition task. Therefore,  we design a noise-based loss named NoiRetrieval Loss, which pays more attention to the learning of low-quality samples and introduce a new setting for low-quality image retrieval, aiming to locate visually similar images to a given query within a database, even when the query images are of poor qualities.
\begin{figure*}[t]
	\centering
	\includegraphics[width=1.0\linewidth]{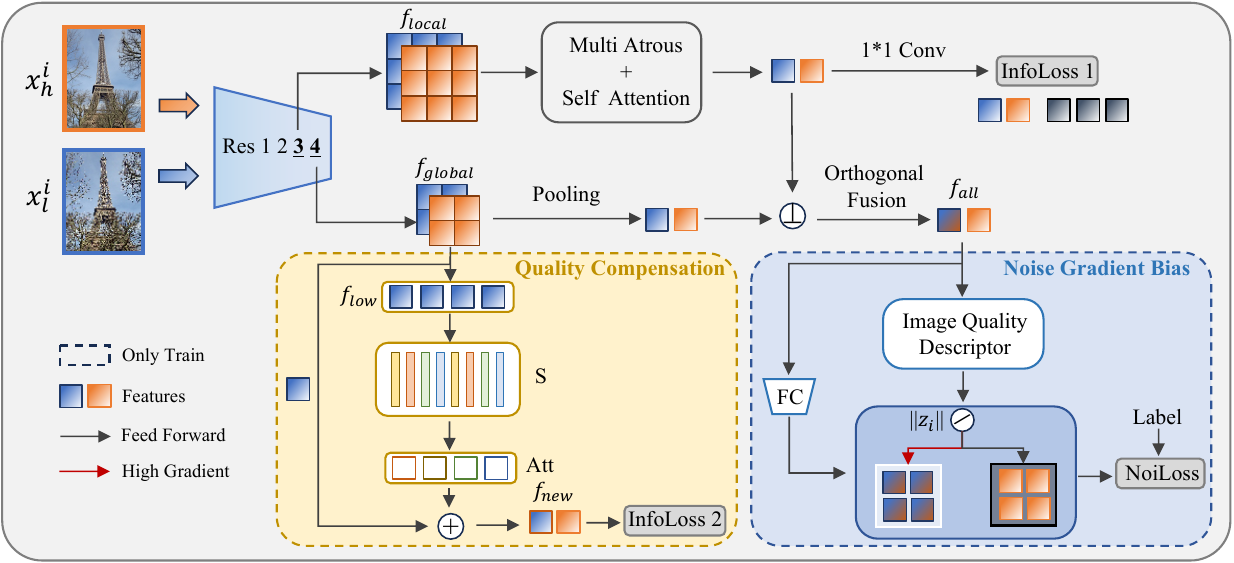}
 \vspace{-0.4cm}
	\caption{The detail structure of our proposed Adaptive Noise-Based Network (AdapNet). Our network is organized into three components: Backbone, Quality Compensation Block (QCB), and Noise Gradient Bias (NGB). The Backbone extracts the local embedding ${f}_{local}$ and global embedding ${f}_{global}$ of input images. The QCB learns the compensation features from the extracted global features of the input image pairs and integrates them with the low-quality features to form ${f}_{new}$. The NGB adjusts the allocation of gradients to the final features ${f}_{all}$ based on image quality $||z_{i}||$, prioritizing the learning of low-quality images.}
\vspace{-0.3cm}
\label{fig:framework}
\end{figure*}

\section{Methods}
\subsection{Problem Setup and Overview}
Recent researches focus on the essential details of an image, thereby enabling efficient image matching. However, the presence of noise, which can originate from natural elements or device-related factors, is often inconsistent across different datasets. This noise can significantly influence the model's image retrieval capabilities~\cite{meng2021magface}. With this challenge in mind, we restructure the conventional image retrieval training input, which typically consists of pairs of low and high-quality images. These paired images are then fed into the network concurrently. Subsequently, we propose a Quality Compensation Block to learn the features associated with the artificially induced noise in the image samples. By emphasizing these noise-specific features, model can effectively differentiate noise from specific image content. In addition to the artificially induced noise, we introduce a Noise Gradient Bias to prioritize the learning of general noise patterns present in low-quality images, enabling the model to handle various types and degrees of unknown noise.

The training datasets include high-quality dataset $\left \{ X_{high}\right \} $ and low-quality dataset $\left \{ X_{low}\right \} $. Then inputs are $N$ image pairs $\left \{ x_{h}^{i},x_{l}^{i}:i\in [1,N], x_{h}^{i}\in X_{high}, x_{l}^{i}\in X_{low}\right \}$. We use $f_{local},f_{global}\in\mathbb{R}^{C\times H\times W}$ as the feature map (channels $C$, height $H$, width $W$) of the input images. $f_{global}$ serves as the input for Quality Compensation Block $\mathcal{S}$ , $f_{local}$ is used to extract local features, and ultimately they are fused into $f_{all}\in \mathbb{R}^{C\ast \times 1\times 1}$ (channels $C\ast$), which is then inputted into the Noise Gradient Bias for Image Quality Descriptor $z$ extraction and classification.

\subsection{Quality Compensation Block}
Since an image pair comes from the same image, the local features are nearly identical. This module aims to learn feature compensation from low-quality to high-quality images through input image pairs. Due to the strong robustness of InfoNCE loss~\cite{oord2018representation}, this compensation is reflected in abstract features. The specific network architecture is depicted in Figure~\ref{fig:framework}. Quality Compensation Block includes eight convolution kernels, used to extract eight different types of noise from the low-quality images. Given the uncertainty of the specific type of noise present in the input image, which could even contain multiple noise types, we selectively fuse multiple compensation features using a $1\times 1$ convolution layer and an adaptiveavgpool2d pooling layer. This fusion process allows the model to adapt to various noise combinations effectively. The entire processing flow of the module can be simplified as follows:
\begin{equation}
    \label{eq:example}  {f}_{new} = \sum_{i=1}^{N} \left( Att\left( \mathcal{S}_i \left( {f}_{low} \right) \right) \right) + {f}_{low}
\end{equation}
where $\mathcal{S}_i$ is employed to represent $N$ different compensation operations and the term $Att$ is used to denote the integration of all compensation features. In Figure~\ref{fig:framework}, the variable ${f}_{global}$ consists of two components: ${f}_{low}$ and ${f}_{high}$, representing the image features of low-quality and high-quality images, respectively. To clarify, we select ${f}_{low}$ as the input for the quality compensation block. The learned compensation features and the original low-quality features are then added together at the element-wise level to obtain new features${f}_{new}$, which are used to combined with the corresponding high-quality image features to compute the InfoNCE loss.
\begin{equation}
    \label{eq:example}{\mathcal{L}}_{info} = -\frac{1}{N} \sum_{i=1}^{N} \log \left( \frac{{\exp(\text{cos}(f_{new}^{i} , f_{high}^{i} ))}}{{\sum_{j=1}^{N} \exp(\text{cos}(f_{new}^{i} , f_{high}^{j} ))}} \right)
\end{equation}
where $N$ signifies the number of low-quality samples. $f_{new}^{i}$ and $f_{high}^{i}$ correspondingly denote the feature vectors of the $ith$ sample and the positive sample. $cos$ represents the similarity function.

\subsection{Noise Gradient Bias}
The cross entropy softmax loss of a sample $x_{i}$ can be formulated as follows,
\begin{equation}
    \label{eq:example}
    l_{CE}(x_{i}) = - log \frac{\exp(W_{y_{i}}z_{i} + b_{i}) }{ {\textstyle \sum_{j=1}^{C}\exp(W_{j}z_{j} + b_{j})}} 
\end{equation}
where $z_{i}\in \mathbb{R}^{d}$ is the feature embedding of $x_{i}$, and $x_{i}$ belongs to the $y_{i}$th class. $W_{j}$ refers to the $j$th column of the last FC layer weight matrix, $W\in \mathbb{R}^{d\times C}$, and $b_{j}$ refers to the corresponding bias term. $C$ refers to the number of classes. To directly optimize the cosine distance in the training objective, ~\cite{liu2017sphereface} use normalized softmax during training. The transformed formula is as follows
\begin{equation}
    \label{eq:example}
   l_{CE}(x_{i}) = - log \frac{\exp(s\cdot \cos \theta_{y_{i}})  }{ {\textstyle \sum_{j=1}^{C}\exp(s\cdot \cos \theta_{j})}} 
\end{equation}
where $\theta_{j}$ corresponds to the angle between $z_i$ and $W_{j}$, $s$ denotes a scale value. In the molecular terms, the component ${\exp(s\cdot \cos \theta_{y_{i}})  }$ is commonly defined as $f(\theta_{y_i},m)$ to enable flexible handling of features across different categories. Based on this, we  design a novel loss function based on image quality different from AdaFace~\cite{kim2022adaface} as follows.

\subsubsection{Image Quality Descriptor}
\begin{wrapfigure}{r}{0.5\textwidth}
    \centering
    \includegraphics[width=1\linewidth]{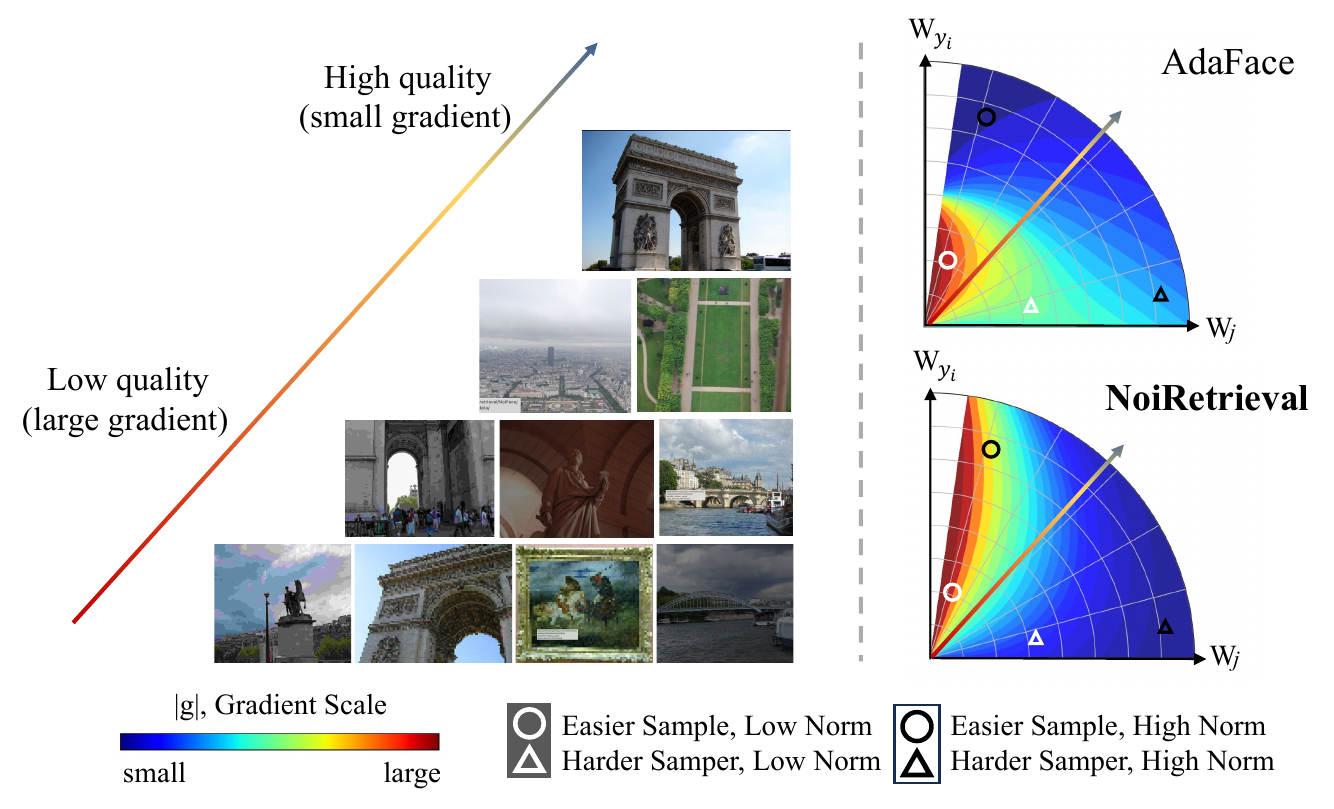}
    \caption{Here we illustrate NoiRetrieval Loss, AdaFace Loss, and their corresponding gradient scaling terms within the feature space. The arc in the feature space represents the angular relationship between a sample and the ground truth class weight vector, $W_{{y}_i}$, as well as the negative class weight vector $W_{j}$. A well-classified sample will be proximate, in terms of angle, to the ground truth class weight vector, $W_{{y}_i}$. Conversely, a misclassified sample will be closer to $W_{j}$. The color within the arc represents the magnitude of the gradient scaling term g. Samples located in the dark red region will contribute more significantly to the learning process.}
	\label{fig:noiretrieval}
\vspace{-10pt}
\end{wrapfigure}
Image quality encompasses various characteristics, including brightness, contrast, motion blur and so on. In the recent AdaFace work, researchers employ the feature norm as a proxy for image quality. It is observed that models trained with a margin-based softmax loss exhibit a correlation between the feature norm and image quality, indicating a obvious trend. Building upon the quality discrimination method used in AdaFace, we make specific modifications to adapt it to a new loss function called NoiRetrieval Loss. The modified Image Quality Descriptor can be formulated as follows:
\begin{equation}
    \label{eq:example}  \widetilde{\left \| {z}_i  \right \|} = \frac{1}{2} \left [  \left \lfloor { \frac{{\left \| {z}_i  \right \|}-{\mu }_z}{{\sigma }_z/h }  }  \right \rceil_{-1}^{1}+1\right ]  
\end{equation}
where $\left \| {z}_i  \right \|$ is the feature norm, ${\mu }_z$ and ${\sigma }_z$ are the mean and standard deviation of all $\left \| {z}_i  \right \|$ within a batch. We introduce the term $h$ to control the concentration and linearly map the original distribution interval from [-1,1] to [0,1] to ensure monotonicity of $\cos z$ in this range.

\subsubsection{NoiRetrieval Loss}
To prioritize the learning of unknown noise patterns typically observed in low-quality images, we design a noise-based loss function according to Image Quality Descriptor. This function emphasizes learning from these images, regardless of whether they are easily recognizable samples or more challenging ones. Specifically,
\begin{equation}
    \label{eq:noiretrieval}  {f({\theta }_j,\widetilde{\left \| {z}_i  \right \|})}_{Noi} = \begin{cases}
 s[cos\widetilde{\left \| {z}_i  \right \|}*cos({\theta }_j + m )] & j={y}_i \\
 s\ cos{{\theta}_j } & j\ne {y}_i
\end{cases}
\end{equation}
where we can observe that when the quality of the image is higher, the value of $\widetilde{\left \| {z}_i  \right \|}$ is larger, which in turn reduce the value of ${f({\theta }_j,\widetilde{\left \| {z}_i  \right \|})}_{Noi}$. This implies that model needs to decrease $\theta_{j}$ to restore the value of ${f({\theta }_j,\widetilde{\left \| {z}_i  \right \|})}_{Noi}$ to its original size, indicating a more strict feature classification for high-quality images. This does not mean the model overlooks low-quality images. On the contrary, during the backpropagation process, it is often the case that gradients from low-quality images are assigned higher weights, rather than at the numerical level when calculating the loss.
Let ${P}^{(i)}_j$ be the probability output at class $j$ after the softmax operation on an input $x_{i}$. By deriving the gradient equations for $L_{CE}$ w.r.t $W_j$ and $x_i$, we obtain the following,
\begin{equation}
{P}^{(i)}_j =\frac{exp(f(cos{\theta }_{{y}_i} ))}{exp(f(cos{\theta }_{{y}_i} ))+{\textstyle \sum_{j\ne{y}_i }^{n}}exp({s\cos {\theta }_j)}}
\end{equation}
we can denote a gradient scaling term (GST)~\cite{kim2022adaface} as
\begin{equation}
\label{eq:gradient}g:=({P}^{(i)}_j-\mathds{1}({y}_i = j))\frac{\partial f(cos{\theta }_j)}{\partial \ cos{\theta }_j}
\end{equation}
since in our model $f(cos{\theta }_{y_i}) =  s[cos\widetilde{\left \| {z}_i  \right \|}*cos({\theta }_j + m )]$ and $\frac{\partial f(cos{\theta }_{y_i}) }{\partial\ cos{\theta }_{y_i}} = s$, the gradient of NoiRetrieval Loss is
\begin{equation}
{g}_{Noi} = {({P}^{i}_j -1 )}cos\widetilde{\left \| {z}_i  \right \|}\left({cos(m) + \frac{cos{\theta }_{{y}_i}sin(m)}{\sqrt{1-cos^{2}{\theta }_{{y}_i} } } }\right)s
\end{equation}
To better illustrate the gradient function, we opt to visually represent it using a heatmap, as shown in Figure~\ref{fig:noiretrieval}. In this heatmap,  we compare the gradients from different dimensions and observe that as the radius increases, which implies a higher quality, the gradient value decreases in both AdaFace~\cite{kim2022adaface} and NoiRetrieval to make the model focus on learning from lower quality images. Moreover, NoiRetrieval tends to allocate more gradients towards simpler samples across different qualities, thereby preserving more sample information, instead of focusing solely on challenging samples as AdaFace~\cite{kim2022adaface} does when image quality is relatively higher. 

Based on the proposed NoiRetrieval Gradient Bias module, we design a novel loss function ${\mathcal{L}}_{Noi} $ which is defined as:
\begin{equation}
    \label{eq:example} {\mathcal{L}}_{Noi} =-\frac{1}{N}\sum_{i=1}^{N}log{\frac{e^{s[cos\widetilde{\left \| {z}_{{y}_i}  \right \|} (cos(\theta +m))] } }{e^{s[cos\widetilde{\left \| {z}_{{y}_i}  \right \|} (cos(\theta +m))] } +B } }  
\end{equation}
where ${y}_i$ is the ground-truth label, $m$ is a margin value, the specific value of which will be provided in the experimental section. $\theta$ represents the angle between the computed feature vector and corresponding label vectors in the mini-batch with size $N$. ${B} = {\textstyle \sum_{j\ne{y}_i }^{n}}e^{s\ cos{\theta}_j } $ is used to calculate the cosine similarity between the current feature vector and other class vectors.

\subsection{Training Objective}
We incorporate the InfoNCE loss into two components of the network, as depicted in the network architecture diagram. $\mathcal{L}_{Info1}$ is employed to ensure the similarity between low-quality and high-quality local features, while $\mathcal{L}_{Info2}$ supervises the learning of distinct compensation features.  Regarding $\mathcal{L}_{Noi}$, it serves to oversee the feature vector that the entire model outputs. Once mapped into the feature space, it assesses the similarity with each category and dynamically adjusts the gradient weights according to the image quality. Hence, it is a primary target for optimization during model training. The final loss function of our proposed network is:
\begin{equation}
    \label{eq:example} \mathcal{L}_{All} = \mathcal{L}_{Noi}+\alpha \mathcal{L}_{Info1} + \beta \mathcal{L}_{Info2}
\end{equation}
where $\alpha$ and $\beta$ are two weights to fuse.
\section{Experiments}
\subsection{Implementation Details}
\subsubsection{Dataset and Evaluation Metric}
We utilize the GLDv2-clean subset of the Google Landmarks dataset v2 ~\cite{weyand2020google} (referred to as GLDv2-clean) as our training dataset. GLDv2-clean comprises 1,580,470 images from 81,313 landmarks, encompassing a wide range of different landmarks. Based on this dataset, we create a new noisy dataset, GLDv2-noisy, by randomly adding various types of noise to the entire dataset using eight manually defined noise functions. During model training, the input images consist of half of the original dataset, while the other half is comprised of the generated noisy dataset. For the evaluation of our model, we utilize the ROxford5k and RParis6k datasets, referred to as ROxf and RPar, respectively. These datasets consist of 70 queries and include 4993 and 6322 database images, respectively. Additionally, we employ the R1M dataset~\cite{radenovic2018revisiting}, which contains one million distractor images, to measure the performance of large-scale retrieval. 
In order to ensure robustness and evaluate the performance of our model under noisy conditions, we introduce random noise to the original datasets, resulting in the Noise ROxford5k and Noise RParis6k datasets, similar to the training dataset. To ensure a fair comparison, we use the mean average precision (mAP) as our evaluation metric on both datasets, following the Easy, Medium and Hard difficulty protocols. Besides, we define another evaluation metric called Noise mAP when query images come from the noisy dataset.

\subsubsection{Training details}
ResNet50 and ResNet101~\cite{he2016deep} are employed as the backbone for conducting experiments in this study. Prior to training, the images are resized to dimensions of $512\times 512$, following the methodology established in previous works~\cite{yokoo2020two,lee2022correlation}. The training process is performed on 8 V100 GPUs for 100 epochs utilizing a batch size of 256. We use the SGD optimizer with a momentum of 0.9. A weight decay factor of 0.0001 is applied, and we adopt the cosine learning rate decay strategy.
\begin{table}[h]
\caption{Result comparisons with baselines on Low-Quality Image Retrieval(\% mAP)  }
\vspace{5pt}
\centering
\resizebox{1\linewidth}{!}{
\begin{tabular}{llcccccccccccc}
\toprule
\multicolumn{2}{c}{\multirow{2}{*}{Method}}&\multicolumn{4}{c}{Easy} &\multicolumn{4}{c}{Medium} &\multicolumn{4}{c}{Hard}\\
\cline{3-6}  \cline{7-10} \cline{11-14} 
\multicolumn{2}{c}{}&Roxf &+1M & Rpar &+1M &Roxf &+1M & Rpar &+1M &Roxf &+1M & Rpar &+1M \\

\toprule
\multicolumn{14}{l}{\textsl{Clear to Clear} } \\ \hdashline
\multicolumn{2}{l}{AdaFace~\cite{kim2022adaface}}&87.2 &77.1 &91.5 &83.8 &65.5 &54.1 &79.9 &63.5 &35.1 &22.3 &60.7 &36.8\\
\multicolumn{2}{l}{CVNet~\cite{lee2022correlation}}&81.5 &68.2 &92.4 &80.2 &59.1 &44.7 &80.3 &56.9 &28.2 &15.8 &60.3 &27.8
\\
\multicolumn{2}{l}{CFCD~\cite{zhu2023coarse}}&82.3 &68.6 &93.2 &82.7 &60.1 &45.8 &80.6 &60.0 &26.2 &14.9 &61.6 &30.7 \\
\multicolumn{2}{l}{SENet~\cite{lee2023revisiting}}&77.6 &61.1 &91.0 &76.3 &55.1 &39.5 &78.5 &52.4 &21.6 &10.2 &57.9 &22.3\\
\multicolumn{2}{l}{AdapNet(Ours)}&\textbf{87.5} &\textbf{77.3} &\textbf{94.3} &\textbf{87.3} &\textbf{67.1} &\textbf{54.5} &\textbf{85.0} &\textbf{67.8} &\textbf{37.7} &\textbf{24.1} &\textbf{69.4} &\textbf{40.8}\\
\toprule

\multicolumn{10}{l}{\textsl{Noise to Clear} } \\ \hdashline
\multicolumn{2}{l}{AdaFace~\cite{kim2022adaface}}&71.8 &58.6 &81.4 &66.4 &53.1 &38.9 &69.6 &48.6 &25.4 &11.6 &49.6 &26.2 \\ 
\multicolumn{2}{l}{CVNet~\cite{lee2022correlation}}&72.1 &53.5 &86.1 &65.4 &50.9 &34.0 &73.3 &44.9 &21.3 &9.0 &51.2 &20.8 \\
\multicolumn{2}{l}{CFCD~\cite{zhu2023coarse}}&70.2 &48.9 &80.1 &63.0 &49.5 &30.8 &69.4 &44.7 &18.8 &8.0 &48.9 &21.4 \\
\multicolumn{2}{l}{SENet~\cite{lee2023revisiting}} &64.5 &44.4 &82.5 &61.9 &46.2 &28.0 &70.9 &42.0 &17.6 &6.2 &48.9 &16.6 \\
\multicolumn{2}{l}{AdapNet(Ours)}&\textbf{75.4} &\textbf{60.4} &\textbf{86.2} &\textbf{69.1} &\textbf{57.1} &\textbf{41.1} &\textbf{76.3} &\textbf{52.1} &\textbf{30.2} &\textbf{14.4} &\textbf{58.2} &\textbf{29.7} \\
\cline{1-14}
\end{tabular}
}
\vspace{-0.7cm}
\label{tab:noise setting}
\end{table}
For the NoiRetrieval Loss, we empirically set the margin parameter $m$ to 0.15. As for global feature extraction, we produce multi-scale representations as well, using 5 scales, $\left \{ \frac{1}{2\sqrt{2} },\frac{1}{2} ,\frac{1}{\sqrt{2} },1,\sqrt{2}\right \} $, to extract final compact feature vectors. Following previous works~\cite{noh2017large,cao2020unifying}. For each scale independently, an L2 normalization is applied. These normalized features are then average-pooled to produce the final descriptor~\cite{efe2021dfm}. We use two kinds of experimental settings to ensure fair comparisons.

\subsection{Experimental Results}
In Table~\ref{tab:noise setting}, results (\% mAP) of different solutions are obtained following the Easy, Medium and Hard evaluation protocols of Roxf (+1M), Rpar (+1M) and their noisy dataset. The methods mentioned in Table~\ref{tab:noise setting} all use one-tenth of the dataset from ``GLDv2-clean'' and  ``GLDv2-noisy'' as the training dataset. ``Clear to Clear'' represents a scenario where both the query image and the image to be retrieved are of high-quality. Conversely, ``Noise to Clear'' means the query image is a noisy image, while the image to be retrieved is of high-quality. ResNet50 is uniformly utilized for models that necessitate the use of the ResNet architecture, and pre-trained weights are imported. It can be observed that our solution consistently outperforms existing methods.

\subsubsection{Comparison with the state-of-the-art models}

We re-implement CVNet~\cite{lee2022correlation}, CFCD~\cite{zhu2023coarse} and SENet~\cite{lee2023revisiting} in the official configuration. Notably, our method outperforms the CFCD with a gain of up to 6.9\% on Rpar-Medium, 9.3\% on Rpar-Hard in the ``Noise to Clear'' setting and 4.4\% on Rpar-Medium, 7.8\% on Rpar-Hard in the ``Clear to Clear'' setting. Even with the addition of another one million images to the database, AdapNet still outperform CFCD by a large margin, respectively, which demonstrates the effectiveness of our method to robust image retrieval.When compared with CVNet and SENet, the proposed AdapNet exhibits significantly superior performance across both datasets. These results exhibit excellent performance of our framework for low-quality image retrieval.

For fairness, we compare our proposed model with the state-of-the-art image retrieval models over the years under the original settings in Table~\ref{tab:original setting}. We utilized 80\% of the ``GLDv2-clean'' dataset for training, employing ResNet50 and ResNet101 as the foundational architectures. Results (\% mAP) of different solutions are obtained following the Medium and Hard evaluation protocols of Roxf and Rpar. ``$\star$'' indicates that ``GLDv2-clean'' is used and ``$\diamond$'' indicates that ``SfM-120k'' is used as the training dataset. State-of-the-art performances are marked in bold and our results are summarized at the bottom. The underlined numbers represent the best performances. After a training period of 100 epochs, we conducted tests on Roxf and Rpar, as detailed in Table~\ref{tab:original setting}. As can be seen, although our model focus on noisy images learning, it still performs well in traditional image retrieval tasks. 
\begin{table}{}
\caption{Result comparisons with baselines on Image Retrieval(\% mAP)  }
\vspace{5pt}
\centering
\resizebox{0.8\linewidth}{!}{
\begin{tabular}{llcccccccc}
\toprule
\multicolumn{2}{c}{\multirow{2}{*}{Method}} &\multicolumn{4}{c}{Medium} &\multicolumn{4}{c}{Hard}\\
\cline{3-6}  \cline{7-10} 
\multicolumn{2}{c}{} &Roxf &+1M & Rpar &+1M &Roxf &+1M & Rpar &+1M \\

\toprule
\multicolumn{10}{l}{\textsl{(A) Local feature aggregation} } \\ \hdashline
\multicolumn{2}{l}{R101-HOW-VLAD$\star$~\cite{jegou2010aggregating}}&73.5 &60.4 &82.3 &62.6 &51.9 &33.2 &67.0 &41.8 \\
\multicolumn{2}{l}{R101-HOW-ASMK$\star$~\cite{tolias2016image}}&80.4 &70.2 &85.4 &68.8 &62.5 &45.4 &70.8 &45.4
\\
\multicolumn{2}{l}{R50-FIRe-ASMK$\diamond$~\cite{tolias2016image}}&81.8 &66.5 &85.3 &67.6 &61.2 &40.1 &70.0 &42.9 \\
\multicolumn{2}{l}{R50-MDA-ASMK$\diamond$~\cite{tolias2016image}}&81.8 &68.7 &83.3 &64.7 &62.2 &45.3 &66.2 &38.9 \\
\multicolumn{2}{l}{R50-Token$\star$~\cite{wu2022learning}}&80.5 &68.3 &87.6 &73.9 &62.1 &43.4 &73.8 &53.3 \\
\multicolumn{2}{l}{R101-Token$\star$~\cite{wu2022learning}}&\textbf{82.3} &\textbf{70.5} &\textbf{89.3} &\textbf{76.7} &\textbf{66.6} &\textbf{47.3} &\textbf{78.6} &\textbf{55.9} \\
\toprule

\multicolumn{10}{l}{\textsl{(B) Global single-pass} } \\ \hdashline
\multicolumn{2}{l}{R101-GeM+DSM$\diamond$~\cite{cao2020unifying}}&65.3  &47.6 & 77.4&52.8 &39.2 &23.2 &56.2 &25.0 \\ 
\multicolumn{2}{l}{R50-DELG$\star$~\cite{simeoni2019local}}&78.3 &67.2 &85.7 &69.6 &57.9 &43.6 &71.0 &45.7 \\
\multicolumn{2}{l}{R101-DELG$\star$~\cite{simeoni2019local}}&81.2 &69.1 &87.2 &71.5 &64.0 &47.5 &72.8 &48.7 \\
\multicolumn{2}{l}{R50-DOLG$\star$~\cite{yang2021dolg}} &80.0 &70.5 &89.5 &77.9 &60.8 &44.6 &77.5 &57.5 \\
\multicolumn{2}{l}{R101-DOLG$\star$~\cite{yang2021dolg}} &82.0 &72.4 &90.1 &80.2 &63.8 &48.3 &78.2 &61.3 \\
\multicolumn{2}{l}{R50-CVNet-Global~\cite{lee2022correlation}l$\star$}&81.0 &72.6 &88.8 &79.0 &62.1 &50.2 &76.5 &60.2 \\
\multicolumn{2}{l}{R101-CVNet-Global~\cite{lee2022correlation}l$\star$}&80.2 &74.0 &90.3 &80.6 &63.1 &53.7 &79.1 &62.2 \\
\multicolumn{2}{l}{R50-CFCD$\star$~\cite{zhu2023coarse}}&82.4 &73.1 &91.6 &81.6 &65.1 &50.8 &81.7 &62.8 \\
\multicolumn{2}{l}{R101-CFCD$\star$~\cite{zhu2023coarse}}&\textbf{85.2} &74.0 &91.6 &82.8 &\textbf{70.0} &52.8 &81.8 &65.8 \\
\multicolumn{2}{l}{R50-SENet$\star$~\cite{lee2023revisiting}}&81.9 &74.2 &90.0 &79.1 &63.0 &52.0 &78.1 &59.9 \\
\multicolumn{2}{l}{R101-SENet$\star$~\cite{lee2023revisiting}}&82.8 &76.1 &\textbf{91.7} &\textbf{83.6} &66.0 &\textbf{55.7} &\textbf{82.8} &\textbf{67.8} \\
\multicolumn{2}{l}{R50-SpCa$\star$~\cite{zhang2023learning}}&81.6 &73.2 &88.6 &78.2 &61.2 &48.8 &76.2 &60.9 \\
\multicolumn{2}{l}{R101-SpCa$\star$~\cite{zhang2023learning}}&83.2 &\textbf{77.8} &90.6 &79.5 &65.9 &53.3 &80.0 &65.0 \\
\toprule
\multicolumn{10}{l}{\textsl{Ours} } \\ \hdashline
\multicolumn{2}{l}{R50-AdapNet$\star$}&81.4 &71.3 &88.2 &78.9 &63.9 &48.7 &74.5 &59.2 \\
\multicolumn{2}{l}{R101-AdapNet$\star$}&82.2 &72.8 &90.5 &81.9 &64.6 &51.1 &79.6 &64.7 \\
\toprule
\end{tabular}
}
\label{tab:original setting}
\vspace{-15pt}
\end{table}

\subsubsection{Comparison with the face recognition models}
Since AdaFace~\cite{kim2022adaface} method is designed to train better in the presence of unidentifiable images in the training data, we incorporate it into our model for comparison. As can be seen in Table~\ref{tab:noise setting}, AdaFace has a slight advantage over traditional image retrieval methods in scenario ``Noise to Clear'', but there is still a significant gap compared to our model. Specifically,  our method outperforms the AdaFace with a gain of up to 6.7\% on Rpar-Medium, 8.6\% on Rpar-Hard in the ``Noise to Clear'' setting and 5.1\% on Rpar-Medium, 8.7\% on Rpar-Hard in the ``Clear to Clear'' setting.

\subsection{Ablation Studies}
In this section, we conduct a series of ablation experiments using the ResNet50 backbone to empirically validate the components of AdapNet. The methods mentioned in Table~\ref{tab:ablation} use one-tenth of the dataset from ``GLDv2-clean'' and ``GLDv2-noisy'' as the training dataset.

\begin{table}[h]
\vspace{-0.3cm}
\caption{Results (\% mAP \& Noise mAP) of ablation experiments on different modules.}
\vspace{5pt}
\centering
\resizebox{0.9\linewidth}{!}{
\begin{tabular}{lccccccccc}
\toprule
\multicolumn{1}{c}{\multirow{2}{*}{ Method}} &\multicolumn{1}{c}{\multirow{2}{*}{ InfoLoss 1}} &\multicolumn{1}{c}{ Quality} &\multicolumn{3}{c}{ mAP} &\multicolumn{3}{c}{ Noise mAP}\\
\cline{4-6} \cline{7-9}
\multicolumn{2}{c}{}& Compensation & Easy & Medium & Hard & Easy & Medium & Hard\\
\toprule
\multicolumn{1}{l}{ AdaFace (baseline)} &$\times$ & None &91.5 &79.9 &60.7 &81.4 &69.6 &49.6\\
\multicolumn{1}{l}{ AdaFace$\star$} &$\times$ & None &92.6 &81.3 &63.4 &86.2 &74.1 &53.8\\
\toprule
\multicolumn{1}{l}{\multirow{5}{*}{ NoiRetrieval}}&$\times$ & None & 94.0 & 84.2 & 67.8 & 85.4 & 75.3 & 56.5\\
\multicolumn{1}{c}{}&$\checkmark$ & None & 94.0 & 84.4 & 68.4 & 86.1 & 76.2 & 57.7\\
\multicolumn{1}{c}{}&$\times$ & CrossEntropy & 93.9 & 83.7 & 66.1 & 79.5 & 69.5 & 51.3\\
\multicolumn{1}{c}{}&$\times$ & InfoLoss 2 & 94.0 & 83.9 & 67.5 & 86.1 & 75.5 & 57.4\\
\multicolumn{1}{c}{}&$\checkmark$ & InfoLoss 2 & \textbf{94.3} & \textbf{84.9} & \textbf{69.4} & \textbf{86.2} & \textbf{76.3} & \textbf{58.2}\\
\toprule
\end{tabular}}
\vspace{-0.5cm}
\label{tab:ablation}
\end{table}

\subsubsection{Effect of the proposed modules}
Ablation studies are conducted on the proposed modules to demonstrate their effectiveness. Starting with AdaFace as the baseline, we experiment with various modifications. Initially, we only adjust the Image Quality Descriptor, resulting in the creation of the AdaFace$\star$ method, which employs the same image quality descriptor as ours. As demonstrated in Table~\ref{tab:ablation}, this adjustment leads to an approximate 4\% improvement over the baseline in  Noise mAP. Based on this, we design the NoiRetrieval Loss, which yields significant improvement in both low-quality and conventional image retrieval, notably achieving an approximate 7\% increase in the Hard category.
\begin{figure*}[t]
	\centering
	\includegraphics[width=1.0\linewidth]{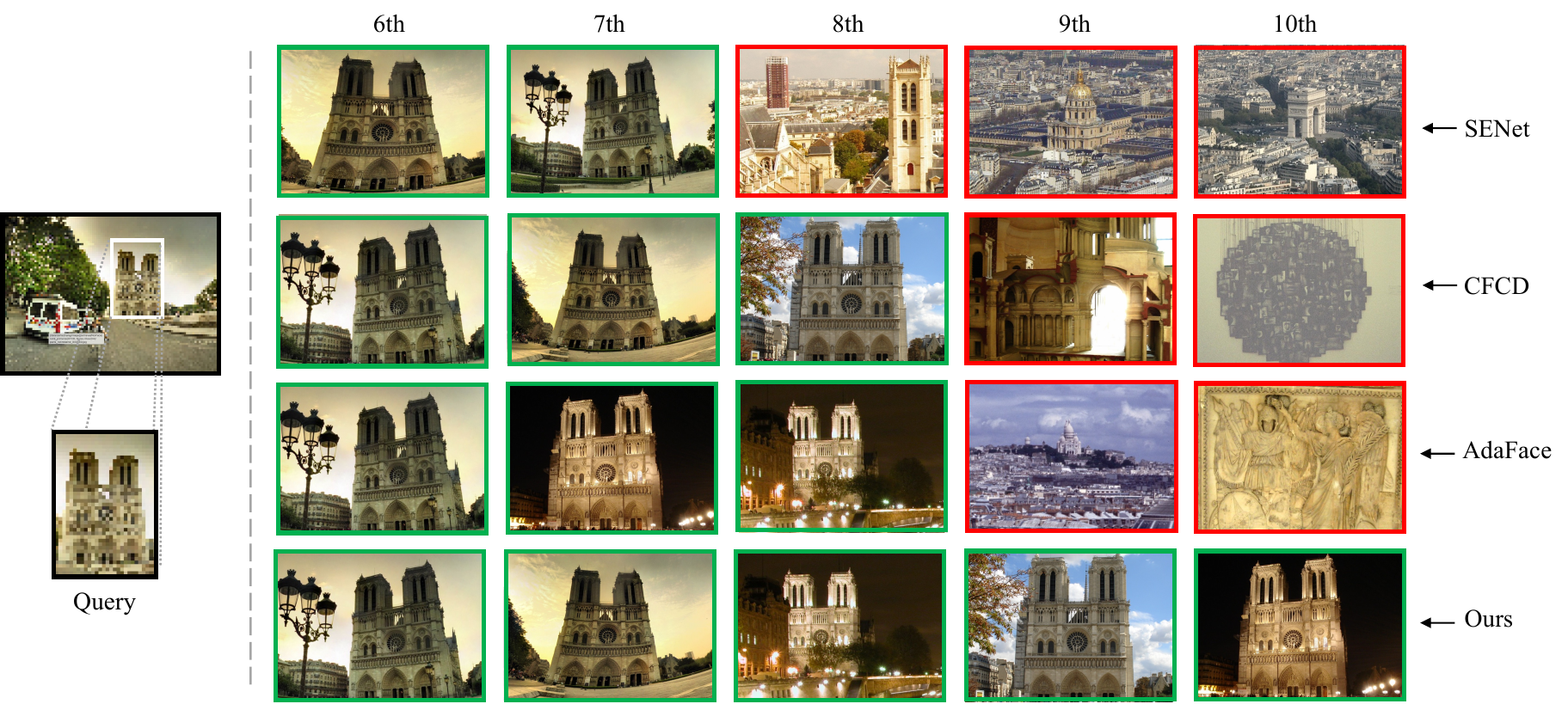}
 \vspace{-0.4cm}
	\caption{The demonstrations of the top retrieved results (ranks 6-10) are shown. The image on the left, used as a query image, is generated by cropping only the part bounded by a white box. On the right, we present the results of CFCD~\cite{zhu2023coarse} and SENet~\cite{lee2023revisiting}  AdaFace~\cite{kim2022adaface}, and our method, displayed from top to bottom. Images enclosed in green and red boxes denote positive and negative images, respectively.}
\vspace{-0.4cm}
\label{fig:result}
\end{figure*}
In subsequent experiments, the NoiRetrieval foundation is augmented with additional modules. CrossEntropy Loss is initially employed in the quality compensation block, but this results in a performance decline. Considering the abstract nature of compensation features, the blurred distinctions between various noise, and the occurrence of multiple noise types within a single image, we choose to employ the more resilient InfoNCE loss. After numerous adjustments and experiments, it is found that using InfoNCE loss to compute similarity loss enhances the model's performance. 

After determining the basic design of the model, we attempt to optimize the model's loss function $\mathcal{L}_{All}$, fixing the coefficient of $\mathcal{L}_{Noi}$ at 1, while $\alpha$ and $\beta$ serve as the hyperparameters. Based on this, we design the following set of experiments, as shown in Table~\ref{tab:parameters}. We find that the results are relatively satisfactory when $\alpha$ and $\beta$ are both set to 0.2.
\begin{table}[h]
\vspace{-0.3cm}
\caption{Results (\% mAP \& Noise mAP) of ablation experiments on different $\alpha$ and $\beta$ in $
\mathcal{L}_{Info}$.}
\vspace{5pt}
\centering
\resizebox{0.8\linewidth}{!}{
\begin{tabular}{ccccccccc}
\toprule
\multicolumn{1}{c}{\multirow{2}{*}{$\mathcal{L}_{Noi}$}} &\multicolumn{2}{c}{$\mathcal{L}_{Info}$} &\multicolumn{3}{c}{mAP} &\multicolumn{3}{c}{Noise mAP}\\
\cline{2-3}\cline{4-6} \cline{7-9}
\multicolumn{1}{c}{} &$\alpha$ &$\beta$ &Easy &Medium & Hard &Easy &Medium & Hard\\
\toprule
1 &0.1 &0.1 &94.3 &84.9 &69.4 &86.2 &76.3 &58.2\\
1 &0.2 &0.2 &94.7 &84.8 &68.7 &87.3 &77.3 &58.4\\
1 &0.2 &0.5 &94.5 &83.9 &67.3 &87.3 &76.6 &57.4\\
1 &0.5 &0.5 &94.3 &84.4 &68.4 &87.8 &77.0 &58.6\\
\toprule
\end{tabular}}
\vspace{-0.8cm}
\label{tab:parameters}
\end{table}

\subsubsection{Qualitative results}
Example qualitative results are shown in Figure~\ref{fig:result}. Despite advanced feature representation, previous solutions that do not consider the quality of images are easily fooled by noise in the dataset. Our proposed network captures known noise and strengthens the learning of unknown noise, thereby enhancing the model's ability to resist noise interference. The retrieval results clearly demonstrate that our model surpasses the other methods when the query image is of low-quality.

\section{Conclusion}
In this work, we introduce a novel setting for low-quality image retrieval and develop an Adaptive Noise-Based Network (AdapNet) to learn robust abstract representations. Specifically, we incorporate a quality compensation block to eliminate known noise from images and devise an innovative adaptive noise-based loss function that dynamically focuses on the gradient relative to image quality, aiming to intensify the learning of unknown noise. Our comprehensive experiments show significant enhancements over state-of-the-art methods on low-quality datasets, while preserving competitive performance on high-quality datasets.

\textbf{Limitations and Broader Impact}. The model relies on the pre-training scheme and the structure of the quality compensation module can be designed more ingeniously. This study enhances image retrieval accuracy in noisy conditions, which can be beneficial for applications involving low-quality images. The proposed datasets and methodologies offer initial insights for future research in handling noisy data in image retrieval tasks.

\bibliographystyle{plain}
\bibliography{neurips_2024}


\end{document}